\def\year{2019}\relax
\documentclass[letterpaper]{article} 
\usepackage{aaai19}  
\usepackage{times}  
\usepackage{helvet}  
\usepackage{courier}  
\usepackage{url}  
\usepackage{graphicx}  
\usepackage{extarrows}

\begin{document}
%
\title{On modelling the emergence of logical thinking}
\author{Cristian Ivan\\
	Romanian Institute of Science and Technology\\
	Cluj-Napoca, Romania\\
	{\tt\small ivan@rist.ro}\\
	{\bf\large Bipin Indurkhya}\\
	Institute of Philosophy\\
	Jagiellonian University\\
	Cracow, Poland\\
	{\tt\small bipin.indurkhya@uj.edu.pl}
}

\maketitle
\begin{abstract}
	Recent progress in machine learning techniques have revived interest in building
	artificial general intelligence using these particular tools. There has been a 
	tremendous success in applying them for narrow intellectual tasks such as pattern 
	recognition, natural language processing and playing Go. The latter application 
	vastly outperforms the strongest human player in recent years. However, these tasks
	are formalized by people in such ways that it has become "easy" for automated 
	recipes to find better solutions than humans do. In the sense of John Searle's 
	Chinese Room Argument, the computer playing Go does not actually understand 
	anything from the game (Linhares and Chada 2013). Thinking like a human mind 
	requires to go beyond the curve fitting paradigm of current systems. There is a
	fundamental limit to what they can achieve currently as only very specific 
	problem formalization can increase their performances in particular tasks. In 
	this paper, we argue than one of the most important aspects of the human mind is 
	its capacity for logical thinking, which gives rise to many intellectual 
	expressions that differentiate us from animal brains. We propose to model the
	emergence of logical thinking based on Piaget's theory of cognitive development.
\end{abstract}

\section{Introduction}
There is much debate about whether mathematics is discovered or invented.
The Platonic view is that mathematical objects exist in a reality separate and independent of ours, with truths we are only discovering. Mathematical truths, e.g. the fact that there is no largest prime number, are independent of ourselves or the existence of the physical universe. We have only found a way to prove it, but the "truth is out there", whether we find a proof or not.

This view is best illustrated in a science fiction novel {\it Contact} by Carl Sagan (1985). Alien intelligence hides meaningful messages in the expansion of $\pi$, the transcendental number that is the ratio of the circumference of a circle to its diameter. The premise of this plot element is that the number $\pi$ is mind-independent, so any intelligence, no matter what form it takes, will be able to find this pattern and decode the message.

The invention camp is based on the idea that mathematics is a cognitive construct, and mathematical objects and theorems reveal as much about our cognitive operations as about the external world. This view is best illustrated by the following long quotation from Jean Piaget: 

\begin{quote}
	``It is agreed that logical and mathematical structures are abstract, whereas physical knowledge - the knowledge based on experience in general - is concrete. But let us ask what logical and mathematical knowledge is abstracted from. There are two possibilities. The first is that, when we act upon an object, our knowledge is derived from the object itself. This is the point of view of empiricism in general, and it is valid in the case of experimental or empirical knowledge for the most part. But there is a second possibility: when we are acting upon an object, we can also take into account the action itself, or operation if you will, since the transformation can be carried out mentally. In this hypothesis the abstraction is drawn not from the object that is acted upon, but from the action itself. It seems to me that this is the basis of logical and mathematical abstraction.
	In cases involving the physical world the abstraction is abstraction from the objects themselves. A child, for instance, can heft objects in his hands and realize that they have different weights --- that usually big things weigh more than little ones, but that sometimes little things weigh more than big ones. All this he finds out experientially, and his knowledge is abstracted from the objects themselves. But I should like to give an example, just as primitive as that one, in which knowledge is abstracted from actions, from the coordination of actions, and not from objects. This example, one we have studied quite thoroughly with many children, was first suggested to me by a mathematician friend who quoted it as the point of departure of his interest in mathematics. When he was a small child, he was counting pebbles one day; he lined them up in a row, counted them from left to right, and got ten . Then, just for fun, he counted them from right to left to see what number he would get, and was astonished that he got ten again . He put the pebbles in a circle and counted them, and once again there were ten. He went around the circle in the other way and got ten again. And no matter how he put the pebbles down, when he counted them, the number came to ten. He discovered here what is known in mathematics as commutativity, that is, the sum is independent of the order. But how did he discover this? Is this commutativity a property of the pebbles? It is true that the pebbles, as it were, let him arrange them in various ways; he could not have done the same thing with drops of water. So in this sense there was a physical aspect to his knowledge. But the order was not in the pebbles; it was he, the subject, who put the pebbles in a line and then in a circle. Moreover, the sum was not in the pebbles themselves; it was he who united them. The knowledge that this future mathematician discovered that day was drawn, then, not from the physical properties of the pebbles, but from the actions that he carried out on the pebbles. This knowledge is what I call logical mathematical knowledge and not physical knowledge.
	The first type of abstraction from objects I shall refer to as simple abstraction, but the second type I shall call reflective abstraction, using this term in a double sense.''
	\cite{piaget1971epistemology}
\end{quote}
(See also 
\cite{dehaene1999number,indurkhya2016,lakatos1976proofs,lakoff2000mathematics,maclane1986mathematics,piaget1971biology}.)

The history of mathematics shows that it is both inspired by natural observations as well as abstract thoughts with no natural correspondence, at least not obvious ones. It is therefore limiting to choose one or the other point of view. Mathematics is both discovered and invented. However, the more abstract a branch of mathematics (or any science relying on a mathematical language) the less obvious it becomes where to draw the line between invention and discovery. 

Looking at mathematics as a human enterprise it becomes clear that it is part of a larger effort of the human mind to understand itself and the world we live in. Rather than describing nature by just enumerating observed objects, a more efficient and effective way is to work with similarities, to observe repeating patterns, to extract invariances. One observed aspect of nature is that it is, to some extent, decomposable in smaller elements which have greater predictive power and generalization capabilities than just lists of objects.
It is exactly this capability of the human mind to observe these patterns and regularities as well as the power to take them in the abstract and operate with them which gives rise to the mathematical science.
Mathematics is rigorous and strict, a type of formalized and  constrained philosophy. It can be seen as an applied branch of philosophy. One could complete, by analogy, the idea of “fields arranged by purity” \cite{xkcd} - illustrated below - and state that ultimately philosophy is just applied human intellect.

\begin{figure}[h]
	\centering
	\includegraphics[width=1\linewidth]{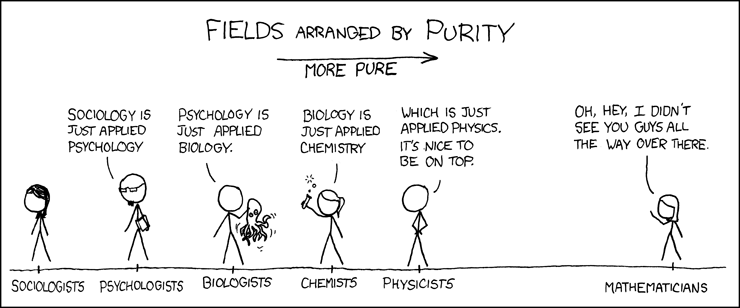}
	\label{fig:purity}
\end{figure}

A rather amusing remark is the fact that for the large majority of Wikipedia's articles by following the first link of the main text and then repeating the process for subsequent articles leads eventually to the ‘philosophy’ article \cite{xefer}. Given the above statement it is only natural for this effect to emerge.

In the light of the above view of the mind, the dilemma often raised in the philosophy of mathematics seem to be a dualistic view which ignores the fact that the human mind is what does the revealing of the mathematical truths - either invented or discovered. It is the human creativity, the ability to infer, reason and argue that lead to those truths. It is therefore more constructive and enlightening to answer another question: where do these abilities of the human mind come from, are they something learned from experiencing the natural world or are they something innate? Paul Erdos, metaphorically, spoke of "The Book" where God wrote the most elegant proofs of mathematical theorems. Some go even further and invoke literal divine inspiration - the possession of truth without proof - as in the case of Srinivasa Ramanujan.

It it only reasonable to argue that the human mind is responsible for both inventions and discoveries, regardless of the field in which it activates. Loosely speaking, invention is the act of putting together concepts which were not associated before. It is a rather technical term and creativity seems to be more broad and less restricted but denotes basically the same idea. Creativity is the free association of any  concepts in any kind of form. It is not restricted by any means and it’s free of any logical burden. Creativity is the essence of art in its many forms stimulating all human senses. It is not bounded by rigour and expresses countless feelings and carries along human emotions and contradictions. 
On the other hand discovery is a more restricted term, it carries rigour and validity, it requires observation, inference, consistency and most of the tools usually attributed to the left side of the brain and it is mostly used in the natural sciences.

In mathematics, invention and discovery go hand in hand and do not exclude one another, quite the contrary. A simple illustration would be the invention of algebraic equations and the subsequent realization that equations of the type $x^{2}=-1$ 
are perfectly valid but which, because of the properties of the multiplication operation, do not make sense and therefore do not have any solution. Human creativity and thinking outside the box led to the invention of complex numbers. In this case, the act of defining complex numbers follows the exact definition of the word ‘invention’. The realization that equations of the previous form exist follows the exact definition of ‘discovery’. It is evident how an abstract invention lead to a discovery which lead to another invention which in turn lead to other discoveries and so on. (See, for instance, \cite{knuth1974surreal,wallace2010infinity}.)

Mathematics can be viewed as creativity under the supervision of logic. It is an interplay between the boundless expansion of creativity and the careful restrictions of rigour, consistency and logic. We can associate together any two or more concepts, but logic will carefully analyze if they can actually live together and not exclude each other. The simple liar's paradox is an example of such seemingly simple association of concepts which, after a careful analysis, leads to an impossibility.

In this context, one can distill two distinct aspects of the mind which come into play: logic and creativity.
Some general remarks on creativity and logic as emergent aspects of the brain are that:

\begin{itemize}
	\item logic appear to be a universal property of healthy brains
	\item logic seems to be common between healthy brains
	\item logic is universal to any mind
	\item creativity is also a property of minds
	\item creativity is as diverse as there are people in the world
	\item mathematics is creativity and logic
	\item mathematics seems to be applicable to nature \cite{1960CPAM...13....1W}
\end{itemize}

Another aspect of the mind which people often use when confronted with insufficiently known situations is intuition. It could be seen in some cases as an insufficiently argued decision process, a guessed, consciously unexplained solution. It is famous for being both extraordinarily useful and correct as well as being the complete opposite. Experience and intuition are strongly linked, the first being, in fact, the source of the latter. It is therefore used as a tool, together with creativity and logic, in the natural sciences, physics in particular being overly abundant in examples of both good and bad intuitions.

The most recent and also most notorious example is the early 20\textsuperscript{th} century physics with the very bizarre aspects of quantum mechanics. Intuition breaks down and, in some artistic sense, human creativity seems to be surpassed by nature's creativity. The only pillar that is not affected by subatomic physics discoveries is logic, although the validity of this statement was debated during the 70s and 80s \cite{Putnam1969} starting from the claim that some basic logical statements are not valid at the level of quantum mechanics and perhaps logic itself should be changed to accommodate the "real rules". Recent analyses debate logic in the context of how to interpret quantum mechanics itself but this extends into metaphysics and shall not be of concern in this work.

\section{Automated minds}

If logic is empirical then it is most probably an emergent property of the nervous system so it is foolish to try to explicitly program it, just like it turned out that human programed rules and features do not stand up to the performance of those discovered by neural networks through “hands-on” experience. Two eloquent examples are:

\begin{itemize}
	\item the explicitly programmed DeepBlue \cite{Campbell:2002:DB:512148.512152} vs. AlphaGo's \cite{alphago} “self play” 
	\item feature engineering in computer vision vs. current convolutional neural networks
\end{itemize}

These are examples of a very particular application of learning through experience but it hints towards the idea that all knowledge available to an automated system should be empirical and not explicitly programmed.

The previous argumentation suggests that a rough approximation of the tools used by the human mind in the intellectual endeavours are creativity, intuition and logic. There is no definite separation as to which is used when, but it is clear that there is a mixture of all of them. In the context of artificial general intelligence, it is important to identify these basic characteristics of an automated system in order to assess its performance. Therefore a few tests would be necessary to be developed as tools to investigate the status of these sub-modules. Important traits of the human mind for which test have to be developed can be summarized as: \textbf{pattern recognition}, \textbf{creativity}, \textbf{intuition} and \textbf{logic}.

Pattern recognition is a field in which we generally started to have some success and spectacular success in very particular domains. However we have just scratched the surface of the domain, though, as many varieties of such tasks are still below human capabilities. 

It is important to understand thoroughly how pattern recognition mechanisms work in the human brain and the animal counterpart because there is a lot of common ground between them. The \textit{creative}, \textit{intuitive} and \textit{logical} cognitive aspects of the brain are less evident in animals and to achieve an artificial general intelligence one would argue that at least these characteristics should be integral parts of any such intelligence as they qualitatively go beyond pattern recognition capabilities.

Logic is an emergent feature of the brain seen at a conscious cognitive level and endowing machines with our prefabricated laws of logic would not be practical nor shine much light on how the brain actually obtains them. The great challenge in the exploration of the human intelligence resides in the difficulty of the investigation tools. Trying to understand how intelligence emerges from the brain via bottom-up biology is like trying to understand how the biological structures emerge from the laws of quantum physics.
Conversely, trying to understand how intelligence emerges from the brain via top-down cognitive science approach is like trying to understand the laws of quantum physics starting from the biological structures of the brain. 

However difficult both of these approaches are, they are the only ones we have and both shine some light into the issue of intelligence. Biology hints towards the mechanism while cognitive science guides the inquirer towards the correct landmarks which have to be achieved in order to approach a general intelligence.

\section{(De)composability}

The world seems to be compositional and procedural. Compositional because we observe that objects in nature are decomposable into smaller objects. Procedural because most things are the result of some process, a sequence of actions/events which lead to the creation of the object/event. There is an apparent hierarchy in which objects at our scale are composed of more elementary objects and other objects can be obtained by combining several different objects at our scale. 

Decomposability is apparent in the fact that mountains are made of large rocks gathered together in one place, and rocks are made of smaller rocks glued together and so on. Composability appears from the fact that smaller rocks can be combined together into bricks which can be combined together to make shelter. The world is a hierarchical two-way structure on which we can go up and down. Does this (de)composability  induce the causal thinking paradigm? Cause then seems to be the decomposition into sequentially structured events which are the only sequence modelling the process by which an event or object comes into being.

The decomposability of the world also leads to the idea that stuff is made of smaller stuff. We observe this to be true everywhere so we have this paradigm hardcoded into our brains, or even worse: hardwired! Having this principle at such a low level in our minds we always think of asking the question of what stuff is made of. What is this rock made of? Smaller rocks. And the smaller rocks are made of even smaller rocks and so on “all the way down”, ending in the same place as the atomists. This naive view has been revised and now we know that rocks are made of crystals, made of atoms, made of electrons and nuclei, made of protons and neutrons made of quarks and gluons made of \dots nothing (yet). So in some sense we are in the same place as the atomists but we like to wrap it up in more elaborate, pretentious academic words.

\begin{figure}[h]
	\centering
	\includegraphics[width=0.7\linewidth]{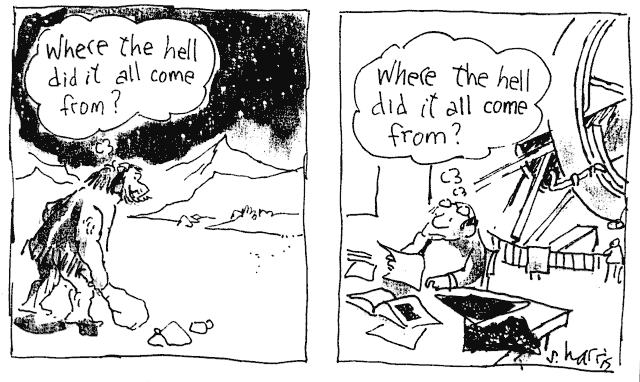}
	\label{fig:wherethehell}
\end{figure}

But the question of what are the (...) made of still persists because of our fundamental view of the world. The situation is so bad that we even start pondering the question of what numbers are made of, in search of more elementary constituents. Since numbers seem to be a made-up concept of real observations the ridiculous of the situation shines through when we realize that we ask the question of what made-up stuff is made of. What are ideas made of? The approach of “what is X made of” seems to lead to an impasse. Picturing this as a hierarchical structure one can realize that there is no reason to believe there is a limit to its height while the lower end seems at least ill defined. We should probably abandon this paradigm and accept that there is a level at which everything has to stop.

\section{Explanation and understanding}

At a classical level where we do not decompose objects and concepts indefinitely and go into metaphysical issues, from a cognitive point of view we observe composability and decomposability down to a certain level after which we do not need to decompose things to explain them, we just observe and experience them. It is impossible to learn the meaning of words by reading their definition from a dictionary and no amount of words and symbols could make us understand the color red. Reading words about the color red does not constitute an explanation and, in fact it is unexplainable, just like many other objects, ideas, feelings etc. do not require explanations but experimentation.

Objects, ideas, concepts, for which decomposition into more elementary components is indeed possible are more suitable to be understood in terms of explanations, unlike elementary objects which require experimentation. It appears, therefore, that explanation means decomposing a new and yet unknown concept into more elementary concepts (known to the agent that needs explanation) and understanding means making the connection between the already known elements and the new concept via the provided decomposition. This is the fundamental aspect of the art of pedagogy, the explainer’s  decomposition of a concept into simpler concepts already known to the explainee. Receiving an explanation in terms of already known notions would sometimes be followed by the exclamation: "It makes sense!". In other words, it would be logical. It is therefore apparent that logical statements are the ones which we already know to be true (because they have been previously  explained) or at least accept as being true due to them being inherently not decomposable into more elementary constituents but just empirically true. 

Decomposability, as discussed in the previous section, applies also to events and processes, not just to objects. Usually events of major importance for us are the result of a sequence of other events. Therefore understanding these sequences is of great importance as well. They are so important, in fact, that they build up the whole reasoning system and lead to concepts of cause and effect. Causal explanations become critical and surpass in importance dry facts exactly because the universe that objectively and subjectively matters to us is procedural and (de)composable, things happen in sequences and the results are very much dependent on what exactly happened and where. This is the reason why we remember facts more easily when they are put in the context of a story, a narrative, rather than being just enumerated. The logical and causal component of the process of understanding is gluing together facts in a more sturdy manner. It is therefore important to understand the process of understanding itself, in both top-down and bottom-up approach as it will shed some light on the emergence of logic.

\section{Ladder of causation}


The ladder of causation from "\textit{The Book of Why}" conceives three qualitatively different levels of intelligence with the first being representative for most animals and the current ML and AI systems. \cite{pearl2018book}.
Indeed artificial neural networks are finding patterns in data, whether they're annotated or not, whether they're collected manually or automatically, association is a tremendously successful application. 

The next two layers in the ladder of causation are based on the question of "what if...". The second layer asks about the future ("what if I do X?" and "how can I make Y?") while the third is concerned about the past ("was X that caused Y?" or "what if I had done X instead of Y?"). 
Although the questions are arranged hierarchically, with the past being placed at a higher level than the future it is easy to see that they have something in common, namely they are both hypothetical questions and are concerned with something that did not happen and is only imagined to happen.
Imagining possible futures requires the same potential as imagining an alternative present.

We can formalize the causal investigation of a system by using the following notation: $\textbf{S}_{t}\xlongrightarrow{\text{do}} \textbf{A} \xlongrightarrow{\text{get   }} \textbf{S}_{t+1}$. We chose the state $\textbf{S}_{t}$ as representing the known present, while the multitude of possible actions $\textbf{A}$ and the multiple possible states $\textbf{S}_{t+1}$ lie in the future and are not known. In the context of the intervention level we have two scenarios. In the first we ask \textit{What if I do...?}, case in which we know we want to pick a certain action, $a_{0}$, and we inquire about the particular state, say $s_{0}$, it will lead us to. The second scenario asks \textit{How can I make Y?}, case in which we know the result we want -- state $s_{0}$ -- but we don't know which particular action to take. In both cases there are two knowns and one unknown.

If we would now move the present to state $\textbf{S}_{t+1}$, put $\textbf{A}$ and $\textbf{S}_{t}$ in the past we would have the same two possibilities of asking the question of an alternative past state given a specific action and present or the hypothetical present given the past state and an alternative action. 

From a cognitive point of view all the above scenarios are similar in nature as they are asking about a hypothetical, alternative state or action in the system. 
The only difference between the two situation is that the future state can be observed after an intervention while the hypothetical past or alternative present can never be observed. However, once the idea of a \textit{What if?} arises it already means a level of intelligence far beyond any pattern recognition machine. It can be classified generally under imagination and creativity, since both these concepts include the idea of something not being there.

The impossibility of answering questions about alternate past states or actions does not shine any light upon a present issue in the absence of a model of the system. Once such a model is present, it starts to bring the two past/future questions and hypotheticals even closer together. Suddenly answers start to fit realistic situations and predictions start to be accurate. The more accurate the model the more accurate the answers. The scientific method is increasingly improving the accuracies of such models applied to the natural world. 
This kind of investigations happen all the time in physics, sometimes the intervention is called an experiment and sometimes it's called a thought experiment. In modern times they are occasionally carried out in computer simulations while in the past scientists like Galileo or Einstein carried them out in their heads, using tools like creativity and logic, the latter being the one judging the correctness of the procedure, the "\textit{gradient descent}" of human reasoning.


Logic guides the inquirer of the "what if" questions towards the correct solution or towards the correct cause of the inquired X or Y. In other words, logic is the fundamental ingredient capable of successfully answer a particular "what if", "how" and "why" question. 

Unless we are capable of building a machine that can perform rudimentary reasoning based on a set of known facts, the dream of achieving at least reliable and casualty free driver-less cars will be eluding us. True artificial intelligence will be one step above that. Artificial general intelligence should be able to learn logical reasoning, not inherit it from its creators.

\section{Architecture}

The recent resounding success of deep learning in various domains have re-sparked the idea and hope that data-driven methods like neural networks will be the key to achieving the more ambitious goal of creating a general artificial intelligence. However,
current implementations of artificial neural networks are confronted with several important issues limiting their capabilities. One major issue with implementing current artificial neural networks (ANN) in practical problems is their rather rigid structure. Networks used for classification tasks have usually fixed input and output sizes. Their internal structure is empirically determined during the training and design of the networks. Convolutional neural networks (CNN), which most closely resemble the human brain hierarchical organization, are chosen such that they best solve the particular given task without clear arguments and reasons as of why the obtained layers/nodes hierarchy structure is optimal for the task. Some task require deeper networks, some require shallower networks, and the optimal network size and structure is found iteratively by the engineers implementing the models. What these networks basically achieve is a decomposition of the input data into simpler structures which, when presented with yet unseen data, are still able to deliver the correct result within the limited scope of the task. Sometimes this decomposition attempts to split the input data into more elements than what is actually needed, and therefore the performance is poorer compared to a rougher decomposition.

Another major limitation of the current ANNs is that they have great difficulties in handling long range correlations. Some of them, like CNNs, are biased to only look for local correlations and we will show in a future work that if the data does not have this property, then they are not capable of producing relevant results. Multilayer perceptrons, on the other hand, which are theoretically capable of finding arbitrarily long correlations, are not computationally feasible and suffer from rigidity as well. Recurrent neural networks are capable of finding longer range correlations and are not as rigid as other feed-forward networks with respect to the input/output sizes but are still not able to tackle arbitrarily long sequences. Deep reinforcement learning techniques have tremendous problems in finding correct correlations between relevant actions and relevant rewards, because in most real world applications the distance between them is generally larger than the correlation capabilities of the networks. Therefore, for some tasks the automated systems still remain far behind the human counterpart.

For the human brain, from a cognitive perspective, one can say that newly learned concepts are further used for building other concepts, ideas, tasks, goals etc. ANNs do not have the possibility to build upon newly acquired skills because of the inflexibility of allocating resources for new situations (and being, in fact, designed for only one particular skill). Dynamic allocation of resources becomes critical. It is not clear how to train a network if its structure is suddenly enriched with a new layer or a few neurons in a layer and still keep the already learned concepts or decompositions. It is known that parts of the brain can take, to some degree, the function of other parts, therefore an artificial network should be capable of performing the same trick, without catastrophically forgetting the previously acquired knowledge. We therefore suggest that a more appropriate neural architecture for either supervised or reinforcement learning should be capable of allocating resources on demand and gradually increase their usage as needed. Letting a neural network dynamically expand to find out by itself the depth/width necessary to decompose certain objects/concepts into more elementary constituents would most likely lead to finding a compact architecture sufficiently complex to satisfactorily complete the given task. 

Experiments show that neural networks used in reinforcement learning (RL) tasks are able to achieve good results when starting from a lower difficulty and gradually increasing it, rather than directly training on the most difficult task. In a sense, gradual increasing of difficulty could be analogous to continuous learning and hierarchical learning from previously acquired skills and concepts. Training networks by gradually increasing the difficulty of their task together with allowing a dynamic growth and allocation of resources would greatly enhance the results obtained from such systems.

One other issue with ANNs today is they are only good at narrow pattern recognition tasks. Some applications, like game playing, have limited success from using end-to-end ML approaches. The recent trend is to employ multiple modules which can perform a narrow task very well and their results are later combined in a predefined system to successfully complete a more complex task. Such a situation is, in some sense, similar to the feature engineering used in earlier versions of machine learning systems.

Current approaches use feature engineering not at the level of data patterns but at the level of patterns in the environment mechanics. The dynamic of the world is decomposed by the engineers into more elementary action units which, combined in various ways, achieve good results for specific tasks. The logic of the world is separated and certain algorithmic responsibilities are given to specialized modules. The burden of finding the problem's logical solution is taken by the engineer and the machine still solves just the pattern recognition and curve fitting task. It becomes clear that one important ingredient is not implemented in the machine yet and it is coming from the outside: logic. The capability of logical reasoning based on data and models is still missing from current systems. 

We will take a brief moment to illustrate the importance of the four basic human traits we have considered in this work -- pattern recognition, creativity, intuition and logic -- when used properly in computer systems. 
One can argue in favour of the AlphaZero \cite{alphazero2018} system that it somehow successfully incorporated all these qualities of intelligence.
The system uses deep learning techniques to select among the more promising future moves at a given board position and these techniques incorporate pattern recognition. After selecting the promising candidates for the next move, a Monte-Carlo tree search algorithm investigates the implications of actually choosing one of the selected moves. One can say that this is analogous to a logical process in the human mind which narrows down the advantages of each move. 
However, due to the large branching factor of the game, one cannot be sure which of the moves is actually the best: only a full map of the game tree could reveal that. Considering that intuition is an insufficiently argued decision process, one can also argue that the selection of the next move is driven by logic as well as intuition. Some professional Go players stated that AlphaGo playing against Lee Sedol, considered the strongest player of our time, has demonstrated that is is capable of creative play. We will leave this as an argument in favor of the claom that AlphaGo and AlphaZero have already achieved creativity.

Despite being so successful, designing systems with built-in logical capabilities is not desirable. This is because they are limited to the domain they are designed for, and also because up to some level, regardless of which side of quantum mechanics we are, logic has to be an emergent quality of the brain. We therefore have to implement a system capable of acquiring logic by itself.

The fundamental reason for the success of current ANN is the geometrical approach to data transformation and the back-propagation algorithm which depends on differentiability of the network components: losses, activation function, convolution operations etc. But this differentiability also constitutes a major impediment in many recent applications of ANN in reinforcement learning approaches because many, if not most, of the real world tasks are not differentiable. In fact the algorithmic world around us behaves more like discrete mathematics. Investigating techniques which go beyond the differentiability paradigm should be a major topic dedicated to the development of neural networks or newer systems.

\section{Synthesis}

We have investigated several aspects of intelligence put in philosophical, cognitive, instrumental and causal contexts. We have shown that in each context logic appears as either an integral part or the essential aspect of intelligence, which distinguishes the human brain from the animal brain. 

We have argued that to go beyond current limited artificial intelligence,logical reasoning is a crucial component of intelligence: it is fundamental to thinking at a lower level than causal thinking, which is a manifestation of the logical part.

Logic is a fundamental aspect of all human philosophic and scientific activities but, as much as it is used, described, defined, taught and talked about, its emergence is not properly understood and investigated. It is exactly this aspect which makes the difference between complex behaviour with a purpose and meaning and one which is mere existential and motivated by survival. Being able to perform any action and thought motivated by logical arguments proved to be the source of the human success in understanding the universe.

The questions of why something is logical, why logic seems to be the same for everyone and why it has so much success when used as a tool for understanding the world, keep eluding us. Efforts for trying to uncover the source of logic should be more focused and practically oriented. Realizing even the smallest logical thinking system without being explicitly programmed for that would be an even greater revolution than ANNs are today, equivalent to the invention of the wheel or the discovery of fire.

\section{Modeling emergence of logical thinking}

To model the emergence of logical thinking, we are essentially taking a Piagetian approach, in which an agent starts with preoperational thinking, proceeds to concrete operational thinking, and then advances to formal thinking \cite{inhelder1958growth}. Though some of Piaget's experiments have been faulted for their design, and the interpretation of their results, we feel that this three-stage model still provides a reasonable framework in which we can address how logical structures emerge from sensori-motor interactions. We are already applying this approach to model computer programming as a cognitive process \cite{perticas2018}. We briefly describe here the three stages:

\begin{description}
	\item[Preoperational thinking:] This embodies the first step towards consolidating the experiences based on sensori-motor explorations. At this stage, symbolic representations emerge, but they are still grounded in static situations. Any transformations are egocentric, in that they are focused on the actions of the agent itself.
	
	\item[Concrete operational thinking:] At this stage, stable representations and systems of transformations emerge: for example, classifications, serial orders, correspondences. These transformations become detached from static situations and become like internalized actions. They also make the agent take a step from the actual to potential, because internalized actions suggest potential outcomes of a situation. This plays a key role in imagination.
	
	\item[Formal thinking:] At this stage, the scope of potentialities become wider so that the reality is considered to be mere one of the possibilities. This corresponds to forming hypotheses and predicting the outcomes of actions. 
\end{description}
In future work, we plan to elaborate this architecture.
The first step is to show how sensori-motor interactions can lead to noticing regularities in the environment, which would form the basis of preoperational thinking \cite{indurkhya_1992_Rq67e4JG}.

\section{Acknowledgements}
This work was supported by the European Regional Development Fund and the Romanian Government through the Competitiveness Operational Programme 2014--2020, project ID P\_37\_679, MySMIS code 103319, contract no. 157/16.12.2016.

\bibliographystyle{aaai}

\end{document}